\documentclass[letterpaper, 10 pt, conference]{ieeeconf}  
\IEEEoverridecommandlockouts
\usepackage{cite}
\usepackage{comment}
\usepackage{balance}
\usepackage[table]{xcolor}
\usepackage{amsmath,amssymb,amsfonts}
\usepackage{subfigure}
\usepackage{graphicx}
\usepackage{comment}
\usepackage{textcomp}
\usepackage{multirow} 
\usepackage{xcolor}
\usepackage{float}
\usepackage{tabularray}
\usepackage{hyperref}
\usepackage{array}
\usepackage{booktabs}
\usepackage{verbatim} 
\usepackage{makecell} 
\usepackage{algorithm}
\usepackage{algpseudocode}
\usepackage{caption}
\usepackage{booktabs}
\usepackage{multirow}
\usepackage{graphicx}
\usepackage{bm} 

\makeatletter
\usepackage[%
    top=0.78in,       
    bottom=0.78in,    
    left=0.78in,      
    right=0.78in      
]{geometry}

\newcommand{\Rmnum}[1]{\expandafter\@slowromancap\romannumeral #1@}

\makeatother
\def\BibTeX{{\rm B\kern-.05em{\sc i\kern-.025em b}\kern-.08em
    T\kern-.1667em\lower.7ex\hbox{E}\kern-.125emX}}

\title{\LARGE \bf Vision-Language Procedural Reasoning for Context-Aware Reward Modeling of Robotic Endovascular Guidewire Navigation}

\author{Wentong Tian$^{1}$, Jiyuan Zhao$^{1}$, Tianliang Yao$^{2}$, Yuxiang Fan$^{1}$, Zhengyu Shi$^{1}$, Dong Liu$^{3}$, Peng Qi$^{1,*}$%
\thanks{This work has been accepted by IEEE/RSJ IROS 2026. Copyright may be transferred without notice, after which this version may no longer be accessible.}
\thanks{This work is supported by the National Key Research and Development Program of China under Grant No. 2023YFB4705200, and the National Natural Science Foundation of China under Grant No. 52575034. The authors would like to thank Mr. Tao Liu from Shanghai Operation Robot Co., Ltd. for providing technical support in experiments. \emph{(*Corresponding Author: Peng Qi, email: pqi@tongji.edu.cn)}}%
\thanks{$^{1}$Department of Control Science and Engineering, College of Electronic and Information Engineering, and Shanghai Institute of Intelligent Science and Technology, Tongji University, Shanghai 200092, China.}%
\thanks{$^{2}$Department of Electronic Engineering, Faculty of Engineering, The Chinese University of Hong Kong, Hong Kong SAR 999077, China.}%
\thanks{$^{3}$Shanghai Operation Robot Co., Ltd., Shanghai 201318, China.}%
}

\begin{document}

\maketitle 
\pagestyle{empty}  
\thispagestyle{empty} 

\begin{abstract}
Robotic-assisted endovascular interventions demand accurate, stable, and context-aware guidewire navigation in complex and patient-specific vascular anatomies. Despite recent advances in robotic precision and learning-based control, existing autonomous navigation methods remain limited by their reliance on static reward functions and the lack of explicit procedural reasoning regarding anatomical context and task progression. To address these challenges, this paper proposes a vision-language procedural reasoning (VL-PR) framework for autonomous guidewire navigation. The framework integrates a multimodal large language model (MLLM) as a procedural reasoning module that interprets real-time visual observations to infer high-level navigation contexts. Instead of generating low-level control commands, the inferred procedural insights enable context-aware reward adaptation by dynamically adjusting the importance of reward components across different navigation phases. This approach allows a single policy to resolve competing objectives and handle complex transitions while preserving a consistent global task goal. Experiments on a physical robotic platform across diverse vascular scenarios demonstrate enhanced task reliability and streamlined navigational efficiency, highlighting the advantages over static-reward methods and offering a scalable solution for complex and multi-task robotic endovascular procedures.
\end{abstract}

\section{Introduction}
Endovascular interventions have become a primary therapy for treating cardiovascular and cerebrovascular diseases. By steering guidewires and catheters through tortuous vasculature, clinicians can reach target lesions with minimal trauma \cite{zhao2026vision, gaudino2023current, liang2026self}. Although clinically effective, precise endovascular navigation remains challenging due to complex vessel geometry and limited visual feedback\cite{pore2023autonomous}. These difficulties result in a steep learning curve and a shortage of experienced clinicians capable of performing such complex procedures \cite{yao2023enhancing, konda2025robotically, yao2025realrecon}. Endovascular robotic platforms have emerged to augment manipulative dexterity and procedural stability, while mitigating operator fatigue and occupational radiation hazard \cite{yao2025advancing,yao2025real}. Nevertheless, the majority of current systems are featured by leader-follower architectures that necessitate persistent low-level manual intervention. Such a human-machine interaction dependency constrains procedural standardization and hinders the scalability of high-precision navigation in complex vascular anatomies \cite{yao2025advancing}.

\begin{figure}[!t]
\centering
\includegraphics[width=0.46\textwidth]{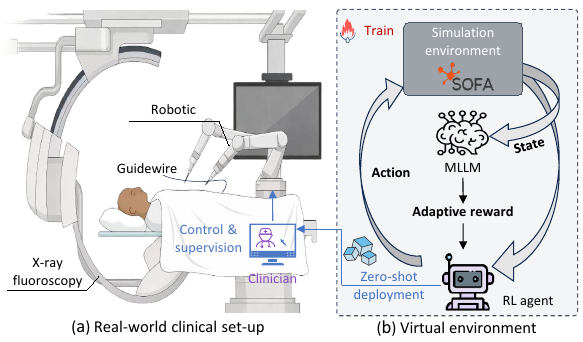}
\caption{Overview of the clinical endovascular intervention scenario and the proposed learning framework. (a) Representative clinical setup of robotic endovascular navigation, where a guidewire is steered through complex vascular anatomies under imaging guidance. (b) Reinforcement learning training pipeline augmented with MLLM semantic reasoning, where high-level contextual understanding derived from vision-language observations informs policy learning. The learned navigation policy is subsequently deployed on the robotic platform without additional policy fine-tuning.} \vspace{-0.6cm}
\label{fig:fig1_overview}
\end{figure}

To advance beyond teleoperation, recent research has prioritized achieving task-level autonomy, aiming to automate discrete procedural segments within the endovascular procedure \cite{yao2025advancing}. Reinforcement learning (RL) has emerged as a robust framework for mapping complex sensory feedback to precise actuation commands. By processing high-dimensional inputs, RL agents learn to execute steering flexible instruments through interaction with the vascular environment \cite{11127627, yao2025sim2real}. Despite its potential, many current RL-based methods for autonomous endovascular navigation still rely on single-policy architectures optimized against static reward structures across the entire procedural continuum. Although adaptive or phase-aware learning has been explored in other surgical domains, such as surgical gesture segmentation and classification \cite{liu2018deep}, procedural reward adaptation remains underexplored for autonomous endovascular guidewire navigation. Such formulations fail to address the heterogeneous, phase-dependent demands of real-world navigation. Specifically, navigating complex vasculatures requires distinct operational modes, such as rapid progression in straight segments and strategic steering at bifurcations. These strategies reflect the clinical intuition of experienced interventionalists, who adapt their approach to address varying anatomical challenges. Existing frameworks lack a mechanism to shift optimization priorities dynamically while maintaining a unified task goal. These limitations necessitate the integration of high-level contextual reasoning to provide stage-aware feedback, thereby enhancing learning stability and robustness during long-horizon endovascular navigation tasks.

MLLMs offer an intuitive framework for procedural reasoning by synthesizing visual observations with linguistic domain knowledge \cite{moor2023foundation}. Recent studies have successfully applied these models to surgical robotics for perception, planning, and decision support. For instance, MLLMs have been utilized for surgical action planning \cite{xu2025surgical, zhang2025csap} and language-guided robotic assistance \cite{11397309}. Building on these advances, endovascular intervention provides an ideal environment for vision-language procedural reasoning due to its unique task demands. Navigation in this domain involves long-duration tasks across diverse anatomical regions where safety requirements are highly context-dependent. Furthermore, recovering from operational difficulties often requires a high-level understanding of the anatomy and interaction risks that exceeds the scope of local geometric features. These characteristics motivate the use of an MLLM to perform procedural reasoning to identify the navigation context. This allows the RL framework to employ context-aware reward adaptation, leading to more robust and intelligent navigation of flexible instruments under a unified task objective.

Leveraging these capabilities, a vision-language procedural reasoning (VL-PR) framework is proposed in this work. This framework utilizes the procedural reasoning of an MLLM to support autonomous guidewire navigation in complex vascular environments. The architecture integrates an MLLM-based procedural reasoning module and an RL agent within a unified pipeline. High-level procedural insights, inferred from real-time vision-language observations, are utilized to condition both policy learning and its execution. Rather than generating direct control commands, the reasoning module provides navigation context to enable context-aware reward adaptation. This design allows the policy to adapt its optimization focus across different navigation phases while maintaining a consistent task goal. The introduction of this context-aware adaptation strategy enables stage-dependent prioritization of control objectives during training, improving convergence efficiency and learning stability in long-duration tasks. A clear separation is maintained between high-level procedural reasoning and low-level motion control, ensuring consistency between training and deployment while improving robustness across diverse vascular anatomies.

The main contributions of this work are summarized as follows:
\begin{itemize}
\item A VL-PR framework for endovascular navigation is established, bridging high-level multimodal reasoning with RL to facilitate intelligent decision-making across complex vascular anatomies.
\item A context-aware reward adaptation strategy is introduced to dynamically prioritize competing objectives, such as balancing traversal efficiency in straight segments with safety-critical precision at bifurcations.
\item A physical validation on an endovascular robotic system is conducted, demonstrating that procedural reasoning enhances task reliability and navigational efficiency within realistic vascular phantoms and challenging anatomical structures.
\end{itemize}

\begin{figure*}[!htbp] 
\centering 
\includegraphics[width=0.92\textwidth]{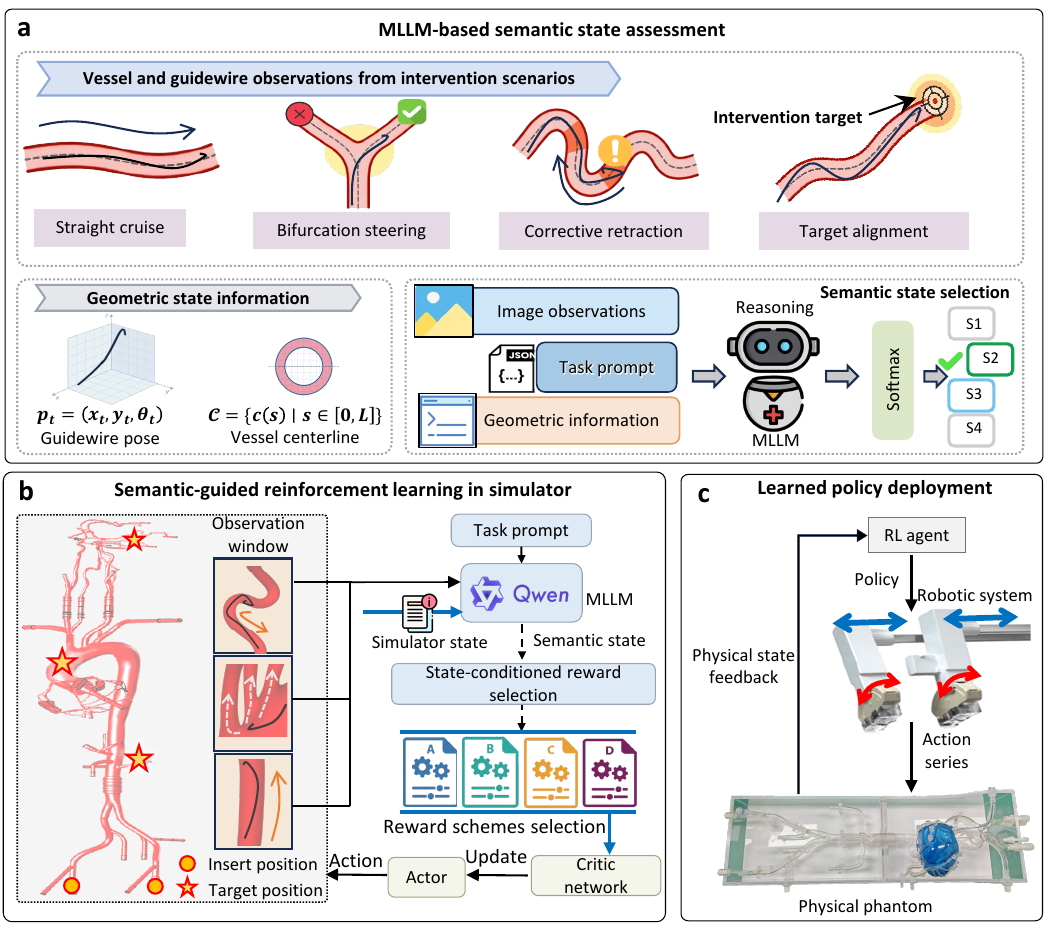} 
\caption{Vision-Language Procedural Reasoning framework for autonomous endovascular navigation. (a) MLLM-based procedural reasoning module. An MLLM processes vision-language observations from the intervention environment to perform procedural reasoning, inferring high-level navigation contexts and task progression. 
(b) Context-aware reward adaptation and policy training. Policy learning is performed in a simulated environment where the MLLM enables context-aware reward adaptation. By dynamically adjusting reward components based on the inferred navigation context, the RL agent learns a robust and goal-directed policy for autonomous guidewire navigation.
(c) Real-world deployment on the physical robotic platform. The learned policy is deployed on the physical robotic system for autonomous guidewire navigation. During execution, the guidewire pose and local vascular geometric features are estimated online by the perception pipeline to provide the state observations required by the policy agent.} 
\label{fig:multi_agent_framework} \vspace{-0.5cm}
\end{figure*}

\section{Methodology}
\subsection{Problem Formulation and Modeling}
Autonomous guidewire navigation is defined as a sequential decision-making task. The objective is to navigate a flexible instrument from an initial insertion point to a designated target vessel within a complex vascular environment while adhering to anatomical and safety constraints. 

At each discrete time step $t$, the system is represented by a state $s_t \in \mathcal{S}$ encoding both the local vascular environment and the guidewire configuration. The state is defined as $s_t = [s_t^{\mathrm{env}}, s_t^{\mathrm{pose}}]$, where $s_t^{\mathrm{env}}$ describes the local vessel geometry and $s_t^{\mathrm{pose}}$ denotes the guidewire tip pose. Based on the current state, a continuous control action $a_t \in \mathcal{A}$ is selected, representing translational and rotational actuation. The execution of $a_t$ results in a transition to the next state $s_{t+1}$ according to the system dynamics. 

The navigation process is modeled as a Markov Decision Process (MDP) defined by the tuple $\mathcal{M} = \langle \mathcal{S}, \mathcal{A}, \mathcal{P}, r, \gamma \rangle,$, where $\mathcal{P}(s_{t+1} \mid s_t, a_t)$ is the state transition distribution, $r$ is the reward function, and $\gamma \in (0,1]$ is the discount factor. The goal is to optimize a policy $\pi_\theta(a_t \mid s_t)$ that maximizes the expected cumulative return
\begin{equation}
J(\pi_\theta) = \mathbb{E}_{\pi_\theta, \mathcal{P}} \left[ \sum_{t=0}^{T} \gamma^t r(s_t, a_t, s_{t+1}) \right],
\label{eq:objective}
\end{equation}
where $T$ represents the episode horizon. 

The reward function $r$ incorporates multiple objectives, including progress toward the target, maintenance of safe vessel wall clearance, and stabilization of guidewire motion. These objectives involve heterogeneous and potentially competing criteria across different navigation phases. This complexity necessitates the introduction of procedural reasoning to infer the navigation context, enabling context-aware reward adaptation to dynamically balance control priorities during the intervention.

\subsection{Procedural Reasoning via Multimodal Large Language Model}
To provide high-level contextual understanding for long-horizon guidewire navigation, a task-specific procedural reasoning module is introduced based on an MLLM. The core objective is to represent the real-time navigation situation as a structured multimodal context and utilize the MLLM to infer procedural states that characterize the anatomical phase and safety-critical conditions.

At each context-update step $t$, the MLLM receives a multimodal observation $o_t=\{ I_t, s_t^{\mathrm{pose}}, g_t^{\mathrm{goal}}, s_t^{\mathrm{geo}} \}$, where $I_t$ denotes the local vascular observation, $s_t^{\mathrm{pose}}$ represents the guidewire tip configuration, and $s_t^{\mathrm{geo}}$ provides local vessel geometry and topology cues. These cues, including centerline direction, local curvature, and bifurcation structures, describe the anatomical context essential for procedural reasoning.

Given $o_t$, the MLLM produces a latent representation $z_t=f_{\mathrm{MLLM}}(o_t;\psi)$, where $\psi$ denotes the model parameters. A discrete procedural state space is defined as $\mathcal{G}=\{\textit{cruise}, \textit{bifurcation},\textit{retraction}, \textit{final\_align}\}$. Each state corresponds to a distinct navigation regime associated with specific control priorities. In particular, the \textit{retraction} state indicates safety-aware withdrawal behaviors required under wall-contact risks, ambiguous progression, or failure recovery. The procedural distribution is computed by a prediction head as
\begin{equation}
p_\psi(g_t \mid o_t) = \mathrm{Softmax}(W z_t + b),
\label{eq:procedural_posterior}
\end{equation}
where $W$ and $b$ are learnable parameters.

To adapt the reasoning module to the endovascular task, a dataset of labeled pairs $(o_t, g_t^\ast)$ is constructed. The procedural labels $g_t^\ast$ are generated using expert-defined heuristics and verified by an operator based on anatomical structure and safety-related events. The primary supervision objective is defined by a categorical cross-entropy loss:
\begin{equation}
\mathcal{L}_{\mathrm{proc}} = -\mathbb{E}_{(o_t, g_t^\ast)} \left[ \log p_\psi(g_t^\ast \mid o_t) \right].
\label{eq:loss_proc}
\end{equation}

Since navigation states exhibit temporal continuity, a temporal consistency regularization term is introduced to encourage coherent predictions and suppress spurious fluctuations:
\begin{equation}
\mathcal{L}_{\mathrm{temp}} = \mathbb{E}_{t} \left[ \left\| z_{t+1} - z_t \right\|_2^2 \right].
\label{eq:loss_temp}
\end{equation}

To prevent overconfident predictions in ambiguous anatomical regions, an entropy regularization term is further incorporated:
\begin{equation}
\mathcal{L}_{\mathrm{ent}} = \mathbb{E}_{t} \left[ - \sum_{g \in \mathcal{G}} p_\psi(g \mid o_t)\log p_\psi(g \mid o_t) \right].
\label{eq:loss_ent}
\end{equation}

The complete training objective for the procedural reasoning module is defined as
\begin{equation}
\mathcal{L}_{\mathrm{MLLM}} = \lambda_1 \mathcal{L}_{\mathrm{proc}} + \lambda_2 \mathcal{L}_{\mathrm{temp}} + \lambda_3 \mathcal{L}_{\mathrm{ent}},
\label{eq:loss_mllm}
\end{equation}
where $\lambda_1, \lambda_2, \lambda_3$ are hyperparameters controlling the relative importance of each term. 

During RL and real-world deployment, the MLLM does not generate direct control actions. Instead, the inferred procedural insights $p_\psi(g_t \mid o_t)$ influence the policy through the context-aware reward adaptation mechanism.

\subsection{Context-Aware Reward Adaptation and Reinforcement Learning}

The procedural context inferred by the reasoning module is integrated into the RL framework by modulating the optimization objective. Instead of switching policies or augmenting the policy input with semantic variables, a unified control policy is maintained while the reward priorities are adjusted according to the inferred procedural state.

The instantaneous feedback is decomposed into four interpretable components describing navigation efficiency, safety, geometric alignment, and mechanical stability \cite{wang2025learning}. These components are arranged in a $2\times2$ reward matrix
\begin{equation}
\Phi(s_t,a_t,s_{t+1}) \triangleq
\begin{bmatrix}
\phi_t^{\mathrm{prog}} & \phi_t^{\mathrm{safe}} \\
\phi_t^{\mathrm{dev}}  & \phi_t^{\mathrm{bend}}
\end{bmatrix}.
\end{equation}

The four components correspond to centerline progress ($\phi^{\mathrm{prog}}$), vessel wall contact safety ($\phi^{\mathrm{safe}}$), deviation from the vessel centerline ($\phi^{\mathrm{dev}}$), and guidewire bending stability ($\phi^{\mathrm{bend}}$), respectively.

The reward components are defined through the following $2\times2$ matrix representation
\begin{equation}
\Phi(s_t,a_t,s_{t+1}) =
\begin{bmatrix}
s_{t+1}-s_t & -\tau_{t+1} \\
-e_{t+1} & -\kappa_{t+1}^{2}
\end{bmatrix}.
\end{equation}

Here $s_t$ denotes the projected arc-length progress of the guidewire tip along the target centerline, $e_t$ is the deviation from the centerline, $\kappa_t$ denotes the local curvature of the guidewire, and $\tau_t$ represents the accumulated number of time steps during which vessel-wall contact occurs. Specifically, $\tau_{t+1} = \tau_t + c_{t+1}$, where $c_t \in \{0,1\}$ indicates whether guidewire–vessel wall contact occurs at time step $t$. In practice, the bending term is approximated from the observed guidewire tip trajectory, which provides a reliable proxy for local guidewire deformation.

Different procedural states emphasize different control priorities. This is achieved through a state-dependent $2\times2$ weight matrix
\begin{equation}
W_g(\xi_t)=
\begin{bmatrix}
w_{g}^{\mathrm{prog}}(\xi_t) & w_{g}^{\mathrm{safe}}(\xi_t)\\
w_{g}^{\mathrm{dev}}(\xi_t)  & w_{g}^{\mathrm{bend}}(\xi_t)
\end{bmatrix},
\end{equation}
where $g\in\mathcal{G}$ denotes the procedural state and $\xi_t$ is a scalar risk indicator defined as
\begin{equation}
\xi_t = 1-\exp\!\left[-(\lambda_\tau\tau_t+\lambda_e e_t+\lambda_\kappa \kappa_t^2)\right].
\end{equation}

For the forward navigation phase (\textit{cruise}), efficient advancement is emphasized while safety penalties increase as risk grows:
\begin{equation}
W_{\mathrm{cruise}}(\xi_t)=
\begin{bmatrix}
w_{0}^{\mathrm{prog}}(1-\xi_t) & w_{0}^{\mathrm{safe}}(1+\xi_t)\\
w_{0}^{\mathrm{dev}}           & w_{0}^{\mathrm{bend}}
\end{bmatrix}.
\end{equation}

In bifurcation regions, geometric alignment and steerability become more critical:
\begin{equation}
W_{\mathrm{bifurcation}}(\xi_t)=
\begin{bmatrix}
w_{0}^{\mathrm{prog}}(1-\xi_t) & w_{0}^{\mathrm{safe}}(1+\xi_t)\\
w_{0}^{\mathrm{dev}}(1+\xi_t)  & w_{0}^{\mathrm{bend}}(1+\xi_t)
\end{bmatrix}.
\end{equation}

When a high-risk configuration is detected (\textit{retraction}), the progress reward is suppressed to allow withdrawal actions that reduce contact duration and restore a stable configuration:
\begin{equation}
W_{\mathrm{retraction}}(\xi_t)=
\begin{bmatrix}
0 & w_{0}^{\mathrm{safe}}(1+\xi_t)\\
w_{0}^{\mathrm{dev}}(1+\xi_t) & w_{0}^{\mathrm{bend}}(1+\xi_t)
\end{bmatrix}.
\end{equation}

Near the target vessel (\textit{final\_align}), the priority shifts toward geometric precision and stable positioning:
\begin{equation}
W_{\mathrm{final\_align}}(\xi_t)=
\begin{bmatrix}
w_{0}^{\mathrm{prog}}(1-\xi_t) & w_{0}^{\mathrm{safe}}(1+\xi_t)\\
w_{0}^{\mathrm{dev}}(1+\xi_t)  & w_{0}^{\mathrm{bend}}(1+\xi_t)
\end{bmatrix}.
\end{equation}

The notation $w_0^{(\cdot)}$ is shared across states, while the corresponding base values are state-dependent and are empirically tuned for each procedural context to reflect stage-specific priorities.

Given the procedural distribution predicted by the multimodal reasoning module, the context-aware reward is computed as
\begin{equation}
r_t =
\sum_{g\in\mathcal{G}}
p_\psi(g\mid o_t)
\left\langle
W_g(\xi_t),
\Phi(s_t,a_t,s_{t+1})
\right\rangle ,
\end{equation}
where $\langle\cdot,\cdot\rangle$ denotes the Frobenius inner product.

Policy learning follows the maximum-entropy RL principle and is implemented using Soft Actor-Critic (SAC). The objective is defined as
\begin{equation}
J(\pi_\theta) =
\mathbb{E}\left[
\sum_{t=0}^{T}
\gamma^t
\left(
r_t +
\alpha\mathcal{H}(\pi_\theta(\cdot\mid s_t))
\right)
\right].
\end{equation}

\begin{figure*}[t] 
\centering 
\includegraphics[width=0.92\textwidth]{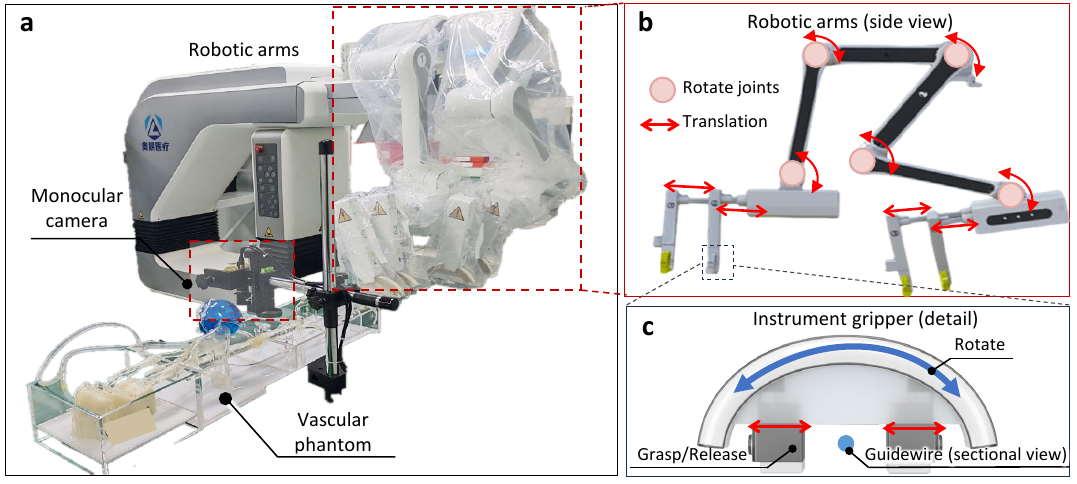} 
\captionsetup{skip=4pt} 
\caption{Experimental robotic platform and setup for phantom-based validation. (a) The robotic intervention system is configured with a realistic vascular phantom to execute procedural navigation tasks. A high-precision actuation unit facilitates guidewire manipulation under the guidance of the VL-PR framework. (b)Illustration of the kinematic structure and degrees of freedom (DOFs) characterizing the robotic manipulator. (c) Detailed view of the gripper mechanism developed for stable guidewire engagement and actuation during autonomous procedures.} \vspace{-0.3cm}
\label{figrob_frame} 
\end{figure*}

\section{Experiments and Results}
\subsection{Experimental Setup}
The experimental setup consists of an anthropomorphic vascular phantom integrated with ALLVAS\texttrademark{} robotic system (Shanghai Operation Robot Co., Ltd., Shanghai, China) and an overhead imaging module (Hikvision, Hangzhou, China). The vascular model replicates complex anatomical features, including branching structures and tortuous segments to provide realistic navigational challenges. Physiological flow dynamics are emulated through a pump-driven circulation loop using a blood-mimicking fluid with controlled viscosity.

Robotic actuation is performed by two coordinated arms equipped with adaptive grippers. These arms operate in a synchronized sequence of anchoring and advancement to manipulate the guidewire. This mechanical design facilitates continuous instrument progression and ensures stable interaction during navigation through complex vascular anatomies. The overhead imaging module captures top-down views mimicking fluoroscopic appearances to support the visual requirements of the navigation system.

\begin{figure*}[!t]
\centering
\includegraphics[width=0.94\textwidth, trim=0cm 9.8cm 0cm 0.7cm, clip]{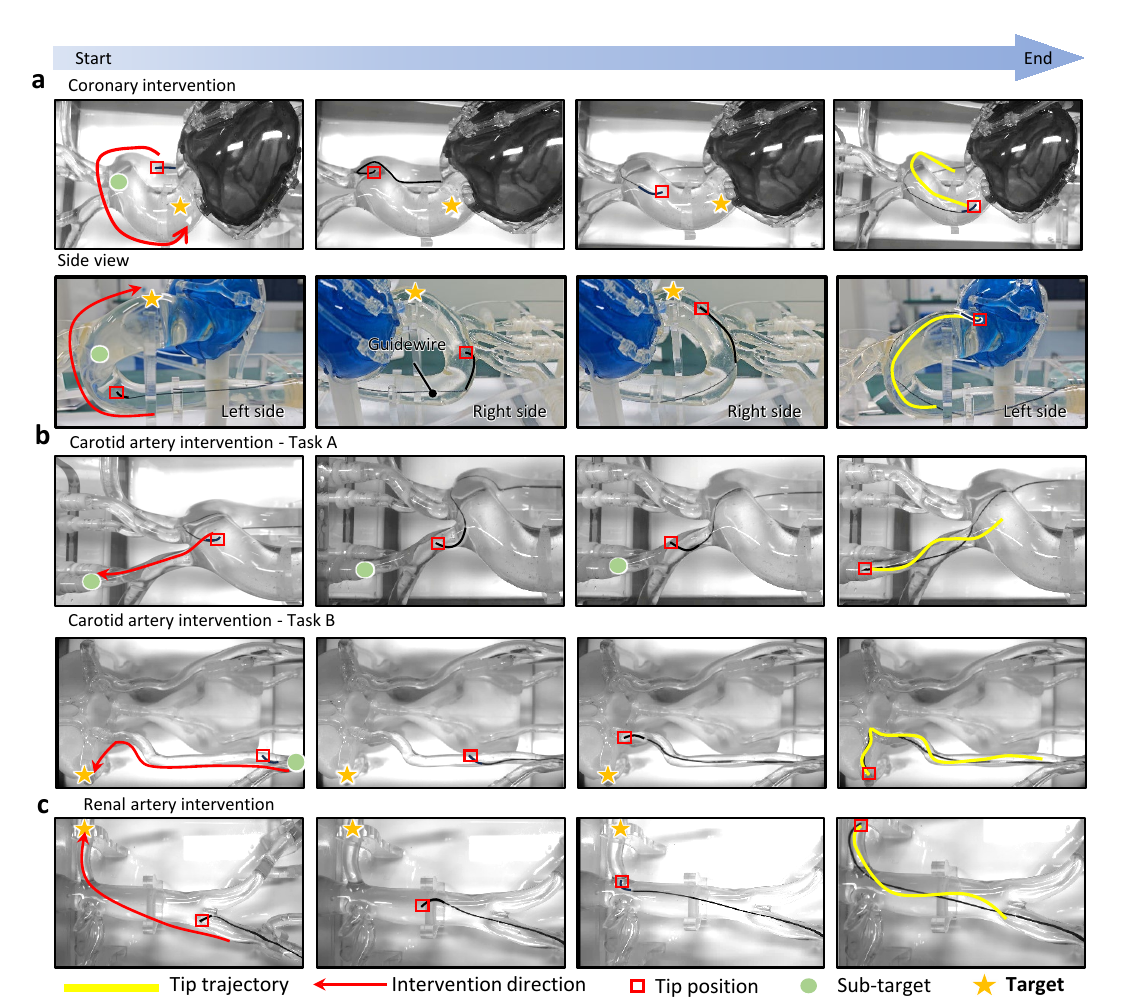}
\caption{Experimental evaluation of the VL-PR framework in diverse vascular scenarios. Representative snapshots illustrate the procedural progression and context-aware navigation across: (a) coronary artery, (b) carotid artery, and (c) renal artery intervention scenarios. For each case, the guidewire trajectory is presented at critical time instants, demonstrating the ability of the proposed framework to interpret navigation contexts and achieve stable, goal-directed movement from initial insertion to the designated target vessel within complex and branching vascular structures.}
\label{fig:exp_results}
\end{figure*}

\begin{table*}[t]
\centering
\caption{Quantitative performance comparison of guidewire navigation across different vascular regions.}
\label{tab:region_comparison}
\renewcommand{\arraystretch}{1.15}
\footnotesize
\begin{tabular*}{\textwidth}{@{\extracolsep{\fill}}l cc cc cc@{}}
\toprule
\multirow{2}{*}{\textbf{Method}} 
& \multicolumn{2}{c}{\textbf{Coronary intervention}} 
& \multicolumn{2}{c}{\textbf{Carotid intervention}} 
& \multicolumn{2}{c}{\textbf{Renal intervention}} \\
\cmidrule(lr){2-3} \cmidrule(lr){4-5} \cmidrule(lr){6-7}
& Success Rate & Steps
& Success Rate & Steps
& Success Rate & Steps \\
\midrule

\textbf{Proposed} 
& \textbf{100\% (30/30)} & \textbf{41.6$\pm$2.1}
& \textbf{97\% (29/30)}  & \textbf{49.3$\pm$2.4}
& \textbf{100\% (30/30)} & \textbf{30.8$\pm$1.8} \\

(w/o MLLM) 
& 73\% (22/30) & 78.5$\pm$3.6
& 77\% (23/30) & 49.7$\pm$3.1
& 73\% (22/30) & 38.4$\pm$2.7 \\

\midrule

TD3 
& 67\% (20/30) & 83.9$\pm$3.7
& 70\% (21/30) & 51.6$\pm$3.1
& 63\% (19/30) & 37.5$\pm$2.8 \\

SAC(standard)
& 70\% (21/30) & 80.6$\pm$3.5
& 73\% (22/30) & 46.4$\pm$3.0
& 73\% (22/30) & 33.2$\pm$2.6 \\

PPO 
& 67\% (20/30) & 99.4$\pm$3.9
& 70\% (21/30) & 59.7$\pm$3.3
& 67\% (20/30) & 41.8$\pm$2.9 \\

DDPG 
& 63\% (19/30) & 112.6$\pm$3.9
& 60\% (18/30) & 67.5$\pm$3.4
& 63\% (19/30) & 44.7$\pm$3.0 \\

Manual 
& 93\% (28/30) & N/A
& 97\% (29/30) & N/A
& 100\% (30/30) & N/A \\

\bottomrule
\end{tabular*}
\end{table*}

\begin{table}[!t]
\centering
\caption{Procedural state classification accuracy comparison between a rule-based baseline and the proposed MLLM.}
\label{tab:state_acc}
\setlength{\tabcolsep}{4.5pt}
\renewcommand{\arraystretch}{1.15}
\footnotesize
\begin{tabular}{lcccc}
\toprule
\textbf{Method} 
& \makecell{\textbf{Cruise}\\\textbf{(N=300)}} 
& \makecell{\textbf{Bifurcation}\\\textbf{(N=170)}} 
& \makecell{\textbf{Retraction}\\\textbf{(N=130)}} 
& \makecell{\textbf{Final\_align}\\\textbf{(N=100)}} \\
\midrule
Rule-based 
& \makecell{80/300\\(26.7\%)} 
& \makecell{42/170\\(24.7\%)} 
& \makecell{36/130\\(27.7\%)} 
& \makecell{22/100\\(22.0\%)} \\
MLLM
& \makecell{\textbf{277/300}\\\textbf{(92.3\%)}} 
& \makecell{\textbf{155/170}\\\textbf{(91.2\%)}} 
& \makecell{\textbf{121/130}\\\textbf{(93.1\%)}} 
& \makecell{\textbf{94/100}\\\textbf{(94.0\%)}} \\
\bottomrule
\end{tabular}
\end{table}

\subsection{Implementation Details}
\label{sec:impl_details}
Experimental validation is conducted on a realistic vascular phantom emulating clinically representative curvatures and bifurcations. A dual-arm robotic system executes an alternating anchoring--advancement mechanism to actuate the guidewire, constituting a closed-loop perception--decision--actuation pipeline for autonomous navigation. A Qwen2.5-VL-3B-Instruct backbone, fine-tuned via 4-bit QLoRA, is integrated as a frozen procedural reasoning module to infer navigation contexts during RL process. The fine-tuning dataset consists of local vascular observations, guidewire tip configurations, and vessel geometry information paired with expert-derived semantic labels. Low-rank adapters are applied to the attention and feed-forward projection layers (rank $r{=}64$, scaling factor $\alpha{=}16$, dropout $0.05$) and optimized using AdamW with a learning rate of $2{\times}10^{-4}$ for three epochs. During RL training and evaluation, the MLLM parameters are fixed and used only to infer the semantic context $g_t$, which influences learning solely through context-conditioned reward modulation. All training is performed on a single workstation equipped with an NVIDIA A100 (80GB) GPU.

The navigational policy is given by a SAC framework with a discount factor $\gamma=0.99$, actor/critic learning rate of $3\times10^{-4}$, batch size of 256, and a replay buffer of size $1\times10^{6}$. Target networks are updated using a soft update rate $\tau=0.005$. The entropy temperature $\alpha$ is automatically tuned with the target entropy set to $-\dim(\mathcal{A})$. Both actor and critic networks are implemented as multilayer perceptrons with two hidden layers of 256 units each, and the critic adopts a Double-Q architecture. To mitigate reward sparsity, a success-biased replay strategy ($p_{\text{succ}}=0.3$) is employed.

\subsection{Cross-Anatomy Evaluation and Procedural Navigation Performance}
The proposed policy is first developed within a high-fidelity, physics-based simulation environment utilizing the SOFA framework. This environment serves as a digital twin of the physical vascular phantom, incorporating identical vessel geometries, deformable behavior, and guidewire-vessel interaction dynamics. Through RL in simulation, a control policy mapping multi-modal observations to continuous manipulation actions is obtained for subsequent real-world deployment. Upon training, the learned policy is deployed on the physical robotic manipulation system for cross-anatomy validation. To facilitate closed-loop execution, a visual perception pipeline is implemented to estimate guidewire states. Monocular camera frames are preprocessed to simulate an X-ray appearance, enhancing contrast for the subsequent modules. Specifically, YOLOv5 is utilized for the detection of the guidewire tip and target regions, while a U-Net architecture performs segmentation to recover the guidewire geometry. This configuration enables the recovery of the guidewire pose required for real-time control. The physical validation is conducted across three anatomically distinct scenarios: coronary, carotid, and renal vasculatures.

Quantitative performance is evaluated against standard continuous-control RL baselines including PPO, DDPG, TD3, and SAC together with expert manual operation. As summarized in Table~I, the proposed VL-PR framework consistently outperforms all baseline RL algorithms across all tested vascular regions. In Coronary and Renal navigation, the proposed method achieves a 100\% success rate, which is substantially higher than SAC with success rates of 70\% and 73\%. In the challenging Carotid scenario, the framework maintains a 97\% success rate, reaching the same level as expert manual operation while demonstrating stronger robustness than conventional RL policies.

Efficiency improvements are also reflected in the reduced navigation steps. For Coronary navigation, the VL-PR framework requires an average of 41.6 steps. Standard SAC requires 80.6 steps, which means the proposed method reduces the required steps by 48.4\%. The lower number of action steps indicates that the reasoning-guided policy generates more direct navigation behaviors within complex vascular structures. Ablation studies reported in Table~\ref{tab:region_comparison} further show that removing the procedural reasoning module leads to a noticeable performance decline. The success rate decreases to a range between 73\% and 77\%, while the number of navigation steps increases significantly. In Coronary navigation the step count increases to 78.5 steps. These results demonstrate that context-aware reward adaptation plays a critical role in resolving competing objectives and enabling reliable navigation in complex branching vascular structures during multi-stage endovascular procedures.

\section{Conclusion and Future Work}
This paper presents a VL-PR framework for autonomous guidewire navigation in endovascular robotic interventions. The proposed approach integrates an MLLM as a procedural reasoning module that interprets real-time vision-language observations to infer high-level navigation contexts. Instead of directly outputting control commands, the inferred procedural insights enable context-aware reward adaptation by dynamically adjusting the optimization objective. This allows a single unified policy to resolve stage-dependent trade-offs among efficiency, safety, and precision. Experiments on a physical vascular platform across coronary, carotid, and renal tasks demonstrate that the proposed method achieves enhanced task reliability and streamlined navigational efficiency compared with conventional learning methods with fixed reward functions, supporting robust autonomous navigation across diverse and complex anatomical structures.

Future research will focus on several directions. First, procedural reasoning will be extended from discrete stage classification to richer structured representations, such as hierarchical graphs or language-based state descriptions. Second, sim-to-real generalization will be strengthened by incorporating domain randomization and vessel deformation variability. Third, safety will be further enhanced by combining context-aware learning with formal safety constraints. Finally, the framework will be evaluated in more clinically realistic settings, including patient-specific vascular anatomies and multi-instrument coordination, to support trustworthy autonomous endovascular robotic interventions.

\bibliographystyle{ieeetr}
\balance
\bibliography{ref}        

@article{gaudino2023current,
  title={Current concepts in coronary artery revascularisation},
  author={Gaudino, Mario and Andreotti, Felicita and Kimura, Takeshi},
  journal={The Lancet},
  volume={401},
  number={10388},
  pages={1611--1628},
  year={2023},
  publisher={Elsevier}
}

@ARTICLE{wang2025learning,
  author={Wang, Haoyu and Tan, Xiaoyu and Yao, Tianliang and Fang, Zhijun and Qi, Peng and Qiu, Xihe},
  journal={IEEE Transactions on Cognitive and Developmental Systems}, 
  title={Learning Expressive Task Embeddings and Sample-Efficient Exploration for Context Shift Reduction in Offline Meta-Reinforcement Environment}, 
  year={2025},
  volume={},
  number={},
  pages={1-15}
}

@INPROCEEDINGS{11127627,
  author={Yao, Tianliang and Ban, Madaoji and Lu, Bo and Pei, Zhiqiang and Qi, Peng},
  booktitle={2025 IEEE International Conference on Robotics and Automation (ICRA)}, 
  title={{Sim4EndoR}: A Reinforcement Learning Centered Simulation Platform for Task Automation of Endovascular Robotics}, 
  year={2025},
  volume={},
  number={},
  pages={824-830},
  doi={10.1109/ICRA55743.2025.11127627}}

@article{konda2025robotically,
  title={Robotically steerable guidewires—Current trends and future directions},
  author={Konda, Revanth and Brumfiel, Timothy A and Bercu, Zachary L and Grossberg, Jonathan A and Desai, Jaydev P},
  journal={Science Robotics},
  volume={10},
  number={105},
  pages={eadt7461},
  year={2025},
  publisher={American Association for the Advancement of Science}
}

@ARTICLE{yao2025sim2real,
  author={Yao, Tianliang and Wang, Haoyu and Lu, Bo and Ge, Jiajia and Pei, Zhiqiang and Kowarschik, Markus and Sun, Lining and Seneviratne, Lakmal and Qi, Peng},
  journal={IEEE Transactions on Automation Science and Engineering}, 
  title={Sim2Real Learning With Domain Randomization for Autonomous Guidewire Navigation in Robotic-Assisted Endovascular Procedures}, 
  year={2025},
  volume={22},
  number={},
  pages={13842-13854}
}

@article{yao2025real,
  title={Real-Time Guidewire Tip Tracking Using a Siamese Network for Image-Guided Endovascular Procedures},
  author={Yao, Tianliang and Pei, Zhiqiang and Li, Yong and Yuan, Yixuan and Qi, Peng},
  journal={Advanced Intelligent Systems},
  volume={7},
  number={10},
  pages={2500425},
  year={2025},
  publisher={Wiley Online Library}
}

@inproceedings{zhao2026vision,
  title={Vision-Based Reasoning with Topology-Encoded Graphs for Anatomical Path Disambiguation in Robot-Assisted Endovascular Navigation},
  author={Zhao, Jiyuan and Shi, Zhengyu and Tian, Wentong and Yao, Tianliang and Liu, Dong and Liu, Tao and Wu, Yizhe and Qi, Peng},
  booktitle={2026 IEEE International Conference on Robotics and Automation (ICRA)},
  pages={21617-21624},
  year={2026},
  organization={IEEE}
}

@inproceedings{yao2025realrecon,
  title={Real-Time {3D} Guidewire Reconstruction from Intraoperative {DSA} Images for Robot-Assisted Endovascular Interventions},
  author={Yao, Tianliang and Li, Bingrui and Lu, Bo and Pei, Zhiqiang and Yuan, Yixuan and Qi, Peng},
  booktitle={2025 IEEE/RSJ International Conference on Intelligent Robots and Systems (IROS)},
  pages={17344--17351},
  year={2025},
  organization={IEEE}
}

@article{yao2023enhancing,
  title={Enhancing percutaneous coronary intervention with heuristic path planning and deep-learning-based vascular segmentation},
  author={Yao, Tianliang and Wang, Chengjia and Wang, Xinyi and Li, Xiang and Jiang, Zhaolei and Qi, Peng},
  journal={Computers in Biology and Medicine},
  volume={166},
  pages={107540},
  year={2023},
  publisher={Elsevier}
}

@article{liang2026self,
  title={Self-Supervised X-Ray Coronary Angiography Segmentation with Vessel-Aware Synthesis Learning},
  author={Liang, Shuang and Liu, Zhicheng and Liu, Guangyuan and Yao, Tianliang and Yang, Chunyi and Qi, Peng},
  journal={IEEE Journal of Biomedical and Health Informatics},
  year={2026},
  publisher={IEEE},
  pages={1-11},
}

@ARTICLE{yao2025advancing,
  author={Yao, Tianliang and Lu, Bo and Kowarschik, Markus and Yuan, Yixuan and Zhao, Hubin and Ourselin, Sebastien and Althoefer, Kaspar and Ge, Junbo and Qi, Peng},
  journal={IEEE Reviews in Biomedical Engineering}, 
  title={Advancing Embodied Intelligence in Robotic-Assisted Endovascular Procedures: A Systematic Review of {AI} Solutions}, 
  year={2026},
  volume={19},
  number={},
  pages={248-266}
}

@article{moor2023foundation,
  title={Foundation models for generalist medical artificial intelligence},
  author={Moor, Michael and Banerjee, Oishi and Abad, Zahra Shakeri Hossein and Krumholz, Harlan M and Leskovec, Jure and Topol, Eric J and Rajpurkar, Pranav},
  journal={Nature},
  volume={616},
  number={7956},
  pages={259--265},
  year={2023},
  publisher={Nature Publishing Group UK London}
}

@inproceedings{zhang2025csap,
  title={CSAP-Assist: Instrument-Agent Dialogue Empowered Vision-Language Models for Collaborative Surgical Action Planning},
  author={Zhang, Jie and Xu, Mengya and Wang, Yiwei and Dou, Qi},
  booktitle={International Conference on Medical Image Computing and Computer-Assisted Intervention},
  pages={139--148},
  year={2025},
  organization={Springer}
}

@inproceedings{xu2025surgical,
  title={Surgical action planning with large language models},
  author={Xu, Mengya and Huang, Zhongzhen and Zhang, Jie and Zhang, Xiaofan and Dou, Qi},
  booktitle={International Conference on Medical Image Computing and Computer-Assisted Intervention},
  pages={563--572},
  year={2025},
  organization={Springer}
}

@ARTICLE{11397309,
  author={Low, Chang Han and Wang, Ziyue and Zhang, Tianyi and Zhuo, Zhu and Zeng, Zhitao and Mazomenos, Evangelos B. and Jin, Yueming},
  journal={IEEE Robotics and Automation Letters}, 
  title={SurgRAW: Multi-Agent Workflow with Chain of Thought Reasoning for Robotic Surgical Video Analysis}, 
  year={2026},
  volume={},
  number={},
  pages={1-8}
  }

@inproceedings{liu2018deep,
  title     = {Deep Reinforcement Learning for Surgical Gesture Segmentation and Classification},
  author    = {Liu, Daochang and Jiang, Tingting},
  booktitle = {Medical Image Computing and Computer Assisted Intervention -- MICCAI 2018},
  series    = {Lecture Notes in Computer Science},
  volume    = {11073},
  pages     = {247--255},
  year      = {2018},
  publisher = {Springer},
  doi       = {10.1007/978-3-030-00937-3_29}
}

@article{pore2023autonomous,
  title   = {Autonomous Navigation for Robot-Assisted Intraluminal and Endovascular Procedures: A Systematic Review},
  author  = {Pore, Ameya and Li, Zhen and Dall'Alba, Diego and Hernansanz, Albert and De Momi, Elena and Menciassi, Arianna and Casals Gelpi, Alicia and Dankelman, Jenny and Fiorini, Paolo and Poorten, Emmanuel Vander},
  journal = {IEEE Transactions on Robotics},
  volume  = {39},
  number  = {4},
  pages   = {2529--2548},
  year    = {2023},
  doi     = {10.1109/TRO.2023.3269384}
}

\end{document}